\documentclass[]{fairmeta}
\usepackage{graphicx}
\usepackage{paracol}
\usepackage{enumitem}
\usepackage{pifont}
\newcommand{\cmark}{\ding{51}} 
\newcommand{\xmark}{\ding{55}} 
\usepackage{times}
\usepackage{amsmath}
\usepackage{latexsym}
\usepackage{tabularx}
\usepackage[table]{xcolor}
\definecolor{cmarkgreen}{RGB}{0, 120, 0}
\definecolor{xmarkred}{RGB}{180, 0, 0}
\newcommand{\cmarkgreen}{{\textcolor{cmarkgreen}{\cmark}}}
\newcommand{\xmarkred}{{\textcolor{xmarkred}{\xmark}}}
\usepackage[T1]{fontenc}
\usepackage[utf8]{inputenc}
\usepackage{enumitem}
\usepackage{microtype}
\usepackage{inconsolata}
\usepackage{graphicx}
\usepackage{xcolor}
\usepackage{needspace}
\definecolor{lightblue}{RGB}{220,230,245} 
\definecolor{navyblue}{RGB}{30,50,100}    
\usepackage{adjustbox}
\definecolor{mygreen}{RGB}{152, 199, 189}
\definecolor{myblue}{RGB}{195, 166, 213}
\definecolor{mypink}{RGB}{213, 166, 195}
\newtcblisting{graylisting}{
  listing only,
  colback=white,
  colframe=gray!50,
  listing options={
    basicstyle=\ttfamily\footnotesize,  
    breaklines=true,
    columns=flexible
  },
  arc=2mm,                
  boxrule=0.3pt,
  left=2pt,
  right=2pt,
  top=2pt,
  bottom=2pt,
  enhanced
}
\usepackage{amsmath}
\usepackage{physics}
\newcommand{\BenchmarkName}{SymPyBench}
\newcommand{\BenchmarkLength}{15,045}
\title{\BenchmarkName: A Dynamic Benchmark for Scientific Reasoning with Executable Python Code}

\author{Shima Imani}
\author{Seungwhan Moon}
\author{Adel Ahmadyan}
\author{Lu Zhang}
\author{Kirmani Ahmed}
\author{Babak Damavandi}

\affiliation{Meta Reality Lab}

\abstract{
We introduce, a large-scale synthetic benchmark of 15,045 university-level physics problems (90/10\% train/test split). Each problem is \textit{fully parameterized}, supporting an effectively infinite range of input configurations, and is accompanied by structured, step-by-step reasoning and executable Python code that produces the ground-truth solution for any parameter set. The benchmark contains three question types: MC-Symbolic (multiple-choice with symbolic options), MC-Numerical (multiple-choice with numerical options), and free-form (open-ended responses). These diverse formats test complementary reasoning skills. By leveraging the dynamic, code-driven nature of the benchmark, we introduce three novel evaluation metrics in addition to standard accuracy: Consistency Score, Failure Rate, and Confusion Rate, that quantify variability and uncertainty across problem variants. Experiments with state-of-the-art instruction-tuned language models reveal both strengths and limitations in scientific reasoning, positioning \BenchmarkName~as a foundation for developing more robust and interpretable reasoning systems.

}

\date{\today}
\correspondence{First Author at \email{shimaimani@meta.com}}

\begin{document}

\maketitle

\section{Introduction}
\label{sec:intro}

\begin{figure*}[ht]
    \centering
    \includegraphics[width=12cm]{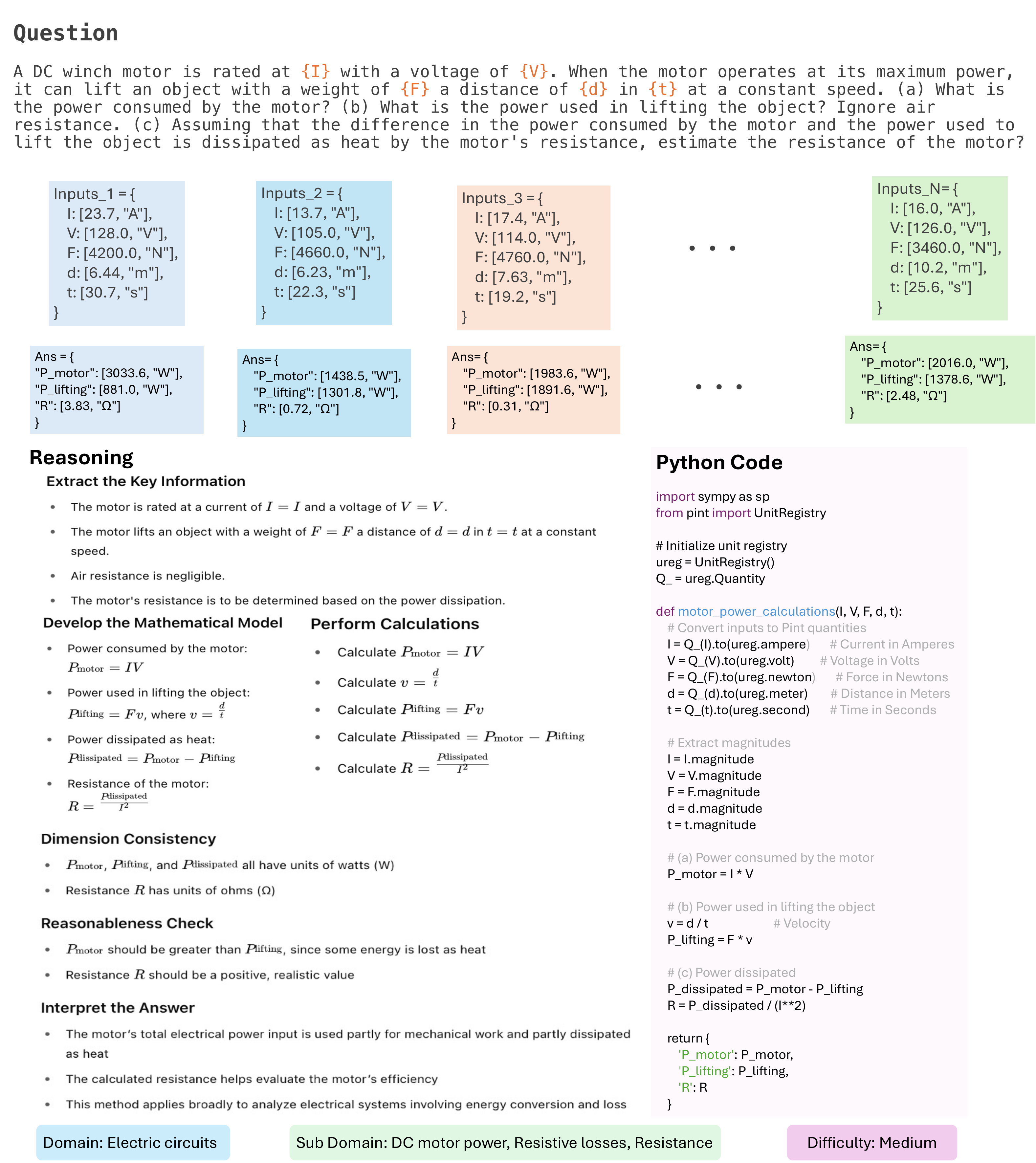}
    \caption{An example from the \textbf{\BenchmarkName}~dataset illustrating a free-form physics question. The figure shows a parameterized problem with variable input parameters, the final answer, detailed step-by-step reasoning, and the associated executable Python code. The question includes metadata such as domain, subdomain, and difficulty.}
    \label{fig:example}
\end{figure*}

Large Language Models (LLMs) have demonstrated impressive capabilities across a wide range of natural language processing tasks~\cite{kojima2022large, claude2024, bai2023qwen, grattafiori2024llama}. Despite this progress, their proficiency in domain-specific, structured reasoning, particularly within scientific disciplines such as physics, remains limited~\cite{ahn2024large, lewkowycz2022solving, chang2024survey}.

Solving physics problems requires the integration of multiple reasoning steps, the precise application of physical laws, and careful mathematical rigor~\cite{larkin1979understanding, hegde2012they, reif1982knowledge}. While existing benchmarks are valuable for evaluating factual recall and fundamental scientific knowledge, they do not fully capture the complexity of structured, step-by-step reasoning that is essential in physics and related domains\footnote{Examples from prior benchmarks in Appendix~\ref{appendix:comparison}.}. Moreover, these benchmarks do not support systematic variation of numerical parameters or linguistic formulations, which limits their ability to effectively evaluate and audit model performance. To address these limitations, we introduce \textbf{\BenchmarkName}, a dynamic benchmark for physics-based reasoning comprising \BenchmarkLength~problem instances paired with executable Python code. Our contributions are:

\paragraph{Dynamic Generalization.}
\BenchmarkName~features systematically parameterized physics problems, where each question can be instantiated with varied input variables. Every instance is accompanied by step-by-step reasoning and executable Python code that produces the corresponding ground-truth solution. The benchmark includes three question types that test complementary reasoning skills. \textit{MC-Symbolic} questions are multiple-choice with symbolic options and primarily evaluate symbolic and algebraic reasoning. \textit{MC-Numerical} questions are multiple-choice with numerical answers, testing a model's ability to perform calculations and apply formulas accurately. \textit{free-form} questions are open-ended and assess the model's ability to generate solutions without any hints, often involving multiple sub-questions or intermediate steps. An example of our benchmark is shown in Figure~\ref{fig:example}.

\paragraph{Beyond Accuracy.} \BenchmarkName~enables systematic evaluation of LLMs through controlled perturbations of problem inputs and linguistic expressions, allowing researchers to probe model behaviors and reveal reasoning patterns. Unlike existing benchmarks that rely on a single problem instance, our dynamic design creates multiple problem variants, enabling a more nuanced assessment of model performance. We introduce novel metrics (\textit{Consistency Score}, \textit{Failure Rate}, and \textit{Confusion Rate}) to capture variability and uncertainty in model reasoning across variants. By analyzing performance across multiple variants, we can determine whether a model consistently applies the correct solution strategy or exhibits inconsistent behavior, failing to generalize across similar problems, thereby providing a more comprehensive understanding of its strengths and weaknesses.

\section{Related Work}
\label{sec:related}

The development of science benchmarks such as ScienceQA~\cite{lu2022learn}, SciBench~\cite{wang2023scibench}, and physics-specific datasets like PhysBench from MMLU~\cite{hendrycks2021measuring} has been instrumental in advancing the evaluation of LLMs on structured reasoning tasks. These benchmarks provide valuable testbeds for assessing baseline scientific knowledge and reasoning skills. Several physics, focused resources, including PhysBench~\cite{hendrycks2021measuring}, SciEval~\cite{sun2024scieval}, and JEEBench~\cite{arora2023have}, primarily adopt multiple-choice formats, which enable standardized evaluation at scale. Many existing benchmarks lack detailed, step-by-step solutions and do not explicitly support symbolic computation, which are essential in scientific disciplines such as physics. Table \ref{tab:benchmark_comparison} summarizes the key differences between \BenchmarkName~and existing scientific reasoning datasets across several dimensions.

While benchmarks are foundational, robust evaluation protocols are equally critical to understand model behavior under variation. Typical evaluations of LLMs often report a single performance metric per dataset, reflecting best-case results under idealized or carefully curated settings. This obscures important dimensions of robustness and reliability~\cite{zhu2024promptbench, bommasani2023holistic}.

PromptBench~\cite{zhu2024promptbench} provides a flexible toolkit for robustness testing, with modules for prompt creation, adversarial generation, and analysis. However, its adversarial prompts can alter input semantics, reducing realism. HELM~\cite{bommasani2023holistic} uses a broader evaluation across metrics like fairness and efficiency, but its robustness tests are limited to minor surface changes and basic contrastive examples~\cite{gardner2020evaluating}.
Building on these efforts, \textbf{\BenchmarkName} introduces a dynamic, parameterized benchmark designed to evaluate model consistency and generalization under controlled variations.

\section{Methodology}
\begin{figure*}[ht]
    \centering
    \includegraphics[width=13cm, height=8cm]{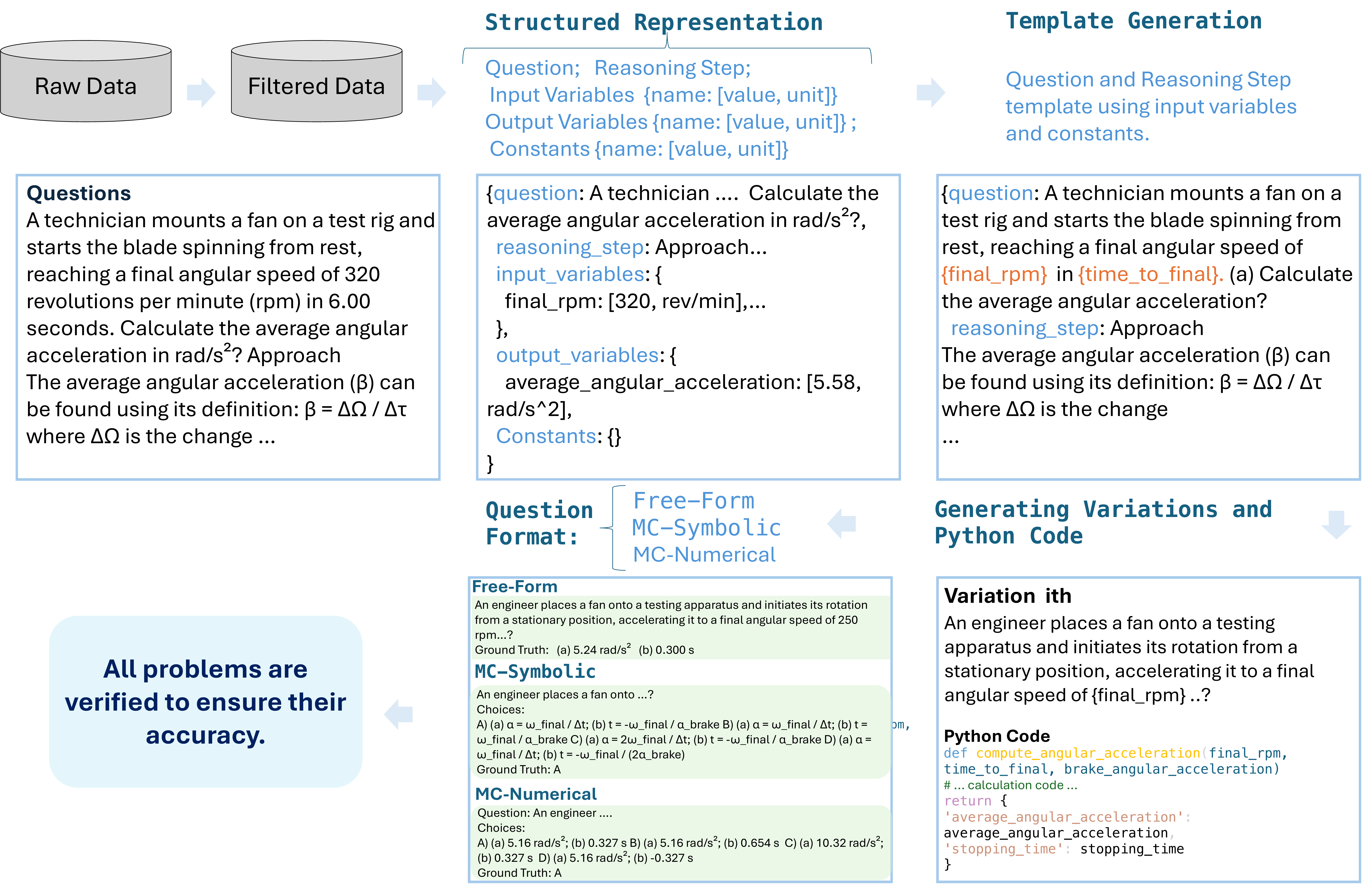}
    \caption{High-level pipeline diagram summarizing the workflow of creating \BenchmarkName.}
    \label{fig:pipeline}
\end{figure*}

We begin by collecting open-source problem sets covering a broad range of undergraduate-level physics topics. The topic distribution reflects the emphasis typically found in standard Bachelor of Science in Physics curricula. All content is sourced from openly available, Creative Commons–licensed materials.\footnote{The dataset is released under a CC-BY-NC license. Portions of the data were generated using LLaMA 3.2 and 3.3 models and are subject to their respective licenses(\url{https://github.com/meta-llama/llama-models/blob/main/models/llama3_2/LICENSE}; \url{https://github.com/meta-llama/llama-models/blob/main/models/llama3_3/LICENSE}).}

\paragraph{Problem Extraction.}
We apply OCR via Tesseract to extract text from each problem and detect any reliance on figures, diagrams, tables, or references to other questions. Dependencies are identified using keyword search and pattern matching. Problems flagged as dependent are excluded, resulting in a dataset of predominantly self-contained, text-based questions suitable for our benchmark. While some dependencies may remain undetected, this step effectively filters most problems that rely on visual or contextual information and prepares the dataset for subsequent processing.

\paragraph{Structured Representation.}
For each problem, we generate a structured textual representation using the LLaMA-3.2-90B-Vision-Instruct model~\cite{grattafiori2024llama}. The model is prompted to reformat the original problem into a standardized schema, as illustrated in Figure~\ref{fig:pipeline}. This process consists of five stages, producing the following components:
\begin{enumerate}[noitemsep, topsep=0pt, leftmargin=*]
\item \textbf{Question:} A clean, self-contained restatement of the problem in natural language.
\item \textbf{Reasoning Step:} A detailed, step-by-step textual explanation outlining the relevant physical principles and intermediate computations.
\item \textbf{Input Variables:} Numerical quantities with their associated physical units (e.g., \texttt{$v1$: [2, m/s]}).
\item \textbf{Output Variables:} Target quantities with their expected final values and units (e.g., \texttt{$v_{a}$: [-3, m/s]}).
\item \textbf{Constants:} Known physical constants such as gravitational acceleration (e.g., \texttt{$g$: [9.8, m/s\textsuperscript{2}]}).
\end{enumerate}
At the conclusion of this process, each question is represented in a structured JSON format encompassing these five components. This structured representation serves as the foundation for generating parameterized code and enables systematic variation across problem instances.

\paragraph{Template Generation.} Building on the structured representation, we prompt the LLM to generate parameterized problem templates. Specifically, the model populates both the question text and the step-by-step solution using symbolic placeholders drawn from the \texttt{Input Variables}, \texttt{Output Variables}, and \texttt{Constants} fields (e.g., \texttt{$v_1$}, \texttt{$F$}, \texttt{$g$}). This ensures that every symbolic reference in the reasoning aligns with the corresponding schema entry, maintaining coherence across intermediate steps and enabling consistent substitution of values across problem variations, as illustrated in Figure~\ref{fig:pipeline}.

\paragraph{Generating Variations and Python Code.} This step consists of two phases: generating textual variations and synthesizing executable Python code.

\textit{Textual Variation Generation.} Given the parameterized template from the previous step, we prompt the model to generate three textual variations of each question. These variations diversify the linguistic phrasing while preserving the underlying problem structure and symbolic placeholders, ensuring linguistic diversity in the benchmark.

\textit{Python Code Synthesis.} We then prompt the model to generate executable Python code that solves each problem. The prompt specifies: (1) the function signature, with \texttt{Input Variables} and \texttt{Constants} keys as input parameters, (2) the expected return format, a dictionary mapping \texttt{Output Variables} keys to their computed values, and (3) the \texttt{Reasoning Steps} to guide the solution logic. This structured guidance enables the model to systematically translate the symbolic solution into executable code, as illustrated in Figure~\ref{fig:pipeline}. We employ few-shot prompting with high-quality examples to improve code generation reliability and consistency.

To validate correctness, we execute each generated Python function by substituting the original numerical values and units from the structured representation. If the computed outputs match the expected \texttt{Output Variables} (accounting for numerical tolerance and unit equivalence), we retain the code in our dataset; otherwise, we discard it. By iteratively refining prompts, we are able to generate correct code for approximately 88\% of problems. For the remaining cases, the generated code does not pass our validation tests.

To ensure correctness and dimensional consistency, all generated code relies on well-established tools~\cite{newell1972human, 10.7717/peerj-cs.103, pint-library}. Specifically, Pint enforces unit consistency across all numerical operations, preventing unit-related errors, while SymPy facilitates symbolic algebra, equation solving, and analytical manipulation, enabling precise mathematical handling throughout the benchmark.

\paragraph{Question Format Generation}
To enable robust and diverse evaluation, we generate problems in three distinct formats: free-form, multiple-choice symbolic (MC-Symbolic), and multiple-choice numerical (MC-Numerical). Our dataset comprises 71.52\% free-form questions, 14.24\% MC-Symbolic questions, and 14.24\% MC-Numerical questions. Note that MC-Symbolic and MC-Numerical formats are generated only for problems with a single sub-question, as many problems in our dataset contain multiple sub-questions. For numerical instantiation, we sample random values for all \texttt{Input Variables} with controlled perturbation (typically $\pm$20 to 50\% of the original values) and substitute them into each question.

\textit{Free-Form Questions.}
Free-form questions are directly derived from the textual variations generated in the previous step. In this format, models must produce numerical answers with appropriate units.

\textit{MC-Symbolic Questions.}
For each MC-Symbolic problem, we generate a multiple-choice question to assess symbolic reasoning capabilities. The correct symbolic answer is obtained by prompting the model to generate an algebraic expression that matches the output of the reference Python implementation. To validate correctness, we substitute $N$ random sets of input variables (typically $N=20$) into both the symbolic expression and the Python code, verifying that outputs match across all test cases.

After validating the correct answer, we generate three distractors by prompting the model to produce small, plausible modifications to the correct expression (e.g., sign changes, term omissions, or altered variable combinations). Each distractor is designed to be algebraically similar yet unambiguously incorrect. To mitigate positional bias, we randomize the order of the four answer choices.

\textit{MC-Numerical Questions.}
For MC-Numerical problems, we substitute the sampled numerical values into the MC-Symbolic format, yielding four numerical options.

\paragraph{Dataset Quality}
All problems are manually reviewed to ensure correctness. Table \ref{tab:data_filtering} shows the percentage of error for each step in each step.

\paragraph{Dataset Composition.} \BenchmarkName~features a diverse set of problems with three types of variations: (1) linguistic variation with three distinct phrasings, (2) format variation with three question formats (free-form, MC-Symbolic, MC-Numerical), and (3) numerical variation with theoretically infinite instantiations. In addition, each problem is annotated with relevant keywords, including domain, sub-domain, and difficulty level. The distribution of problems across high-level physics topics is shown in Table~\ref{tab:data-distribution}. More analysis is provided in appendix.

\begin{table*}[ht!]
\centering
\small
\begin{tabular}{lclclc}
\toprule
\textbf{Mechanics} & \textbf{33.80\%} & \textbf{Electricity and Magnetism} & \textbf{26.76\%}  & \textbf{Modern Physics} & \textbf{12.68\%} \\
\midrule
\quad\textit{Kinematics} &  & \quad\textit{Electric Current} &   & \quad\textit{Quantum Mechanics} &  \\
\quad\textit{SUVAT Equations} &  & \quad\textit{Electric Field} &   & \quad\textit{Special Relativity} &   \\
\quad\textit{Projectile Motion} & & \quad\textit{Lorentz Force} &  & \quad\textit{Photon Energy} &   \\
\midrule
\textbf{Thermodynamics} & \textbf{8.45\%} & \textbf{Waves and Oscillations} & \textbf{11.27\%} & \textbf{Optics} & \textbf{7.04\%} \\
\midrule
\quad\textit{Kinetic Theory of Gases } &  & \quad\textit{Wave Motion} &  & \quad\textit{Geometric Optics} & \\
\quad\textit{Ideal Gas Law} &  & \quad\textit{Frequency} &  & \quad\textit{Polarization} & \\
\quad\textit{RMS Speed} &  & \quad\textit{Doppler Effect} &  & \quad\textit{Electromagnetic Waves } & \\
\midrule
\end{tabular}
\caption{Distribution of problems across physics domains and their top three subdomains in SymPyBench.}
\label{tab:data-distribution}
\end{table*}

\section{Experimental Results and Insights Beyond Accuracy}

\begin{table*}[h]
\centering
\caption{Performance Metrics across LLMs on \BenchmarkName. Includes Partial Accuracy, Exact Match Accuracy, Consistency Score, Complete Failure Rate, and Confusion Rate.}
\label{tab:combined_metrics}
\resizebox{\textwidth}{!}{
\begin{tabular}{lcccccc}
\toprule
\textbf{Model} & \textbf{Partial Accuracy}$\uparrow$ & \textbf{Exact Match Accuracy} $\uparrow$ & \textbf{Consistency Score}$\uparrow$ & \textbf{Complete Failure Rate} $\downarrow$& \textbf{Confusion rate} $\downarrow$\\
\midrule
Qwen2.5-7B-Instruct       & 24.26\% & 16.44\% & 5.66\%  & 41.51\% & 15.09\% \\
Qwen2.5-72B-Instruct      & 66.57\% & 61.69\% & 37.74\% & 15.09\% & 11.32\% \\
Llama-3.3-70B-Instruct    & 59.05\% & 54.17\% & 28.30\% & 15.09\% & 7.55\%  \\
Llama3.1-405b-instruct & 42.79\% & {34.45\%} & {17.46\%} & {30.16\%} & {14.29\%}  \\
Llama4-maverick-17b-128e-instruct& 69.92\% & {64.17\%} & {34.92\%} & {9.52\%} & {11.11\%}  \\
Llama4-scout-17b-16e-instruct& 56.49\% & {50.17\%} & {20.63\%} & 14.29\% & {19.05\%}  \\
OpenAI GPT (gpt-4-turbo)  & 60.59\% & 53.73\% & 33.33\% & 18.18\% & 12.12\% \\
Gemini-2.0-Flash          & \textbf{71.43\%} & 64.49\% & 34.38\% & \textbf{9.38\%}  & 12.50\% \\
Anthropic Sonnet-3.7      & 70.81\% & \textbf{65.48\%} & \textbf{42.42\%} & 18.18\% & \textbf{6.06\%}  \\
\bottomrule
\end{tabular}
}
\end{table*}

We evaluate a range of state-of-the-art instruction-tuned LLMs on \BenchmarkName~to assess their scientific reasoning capabilities under dynamic and perturbed conditions. To this end, we measure several key metrics:

\paragraph{\textbf{Exact Match Accuracy:}} The proportion of problems for which the model produces a completely correct end-to-end solution:

\begingroup
\small
\[
\text{Exact Match Accuracy} = \frac{\text{Number of fully correct solutions}}{\text{Total number of problems}}
\]
\endgroup

Many problems in our dataset are composed of multiple subproblems (e.g., parts a, b, c; see Appendix~\ref{distribution sub-questions}). To calculate exact match accuracy, we require the model to correctly solve all parts of a problem.
However, because many problems are subdivided in this way, we also introduce \textbf{Partial Accuracy} as a complementary metric.

\paragraph{\textbf{Partial Accuracy}:} The fraction of subproblems within a structured solution that the model answers correctly.

\begingroup
\small
\[
\text{Partial Accuracy} =
\frac{
    \text{Number of correct subproblems}
}{
    \text{Total number of subproblems}
}
\]
\endgroup

In addition to accuracy, we evaluate the model’s robustness using several complementary metrics:

\paragraph{\textbf{Consistency Score}:} The proportion of problem groups where the model consistently provides the correct answer across all perturbed variants (i.e., versions of each problem modified by numerical, textual, or format changes), reflecting the stability and reliability of the model's performance.
A high consistency score indicates that the model can reliably solve problems even with slight variations, showcasing its generalization ability.

\begingroup
\small
\[
\text{Consistency Score} =
\frac{
    \text{\# groups with all correct variants}
}{
    \text{\# total problem groups}
}
\]
\endgroup

\paragraph{\textbf{Confusion Rate}:} The confusion rate indicates the fraction of problem groups where the model’s accuracy across variants is around 40\%-60\%, reflecting uncertainty in the model’s reasoning. It provides insight into situations where the model may be guessing or uncertain about the correct approach.

\begingroup
\small
\[
\text{Confusion Rate} =
\frac{
    \text{\# groups with $\sim$50\% accuracy}
}{
    \text{\# total problem groups}
}
\]
\endgroup

\paragraph{\textbf{Complete Failure Rate}:} This metric tracks the proportion of problem groups where the model answers all variants incorrectly. A high Complete Failure Rate indicates areas of consistent failure, providing valuable diagnostic information for improving model performance.

\begingroup
\small
\[
\begin{array}{c}
\text{Complete Failure Rate =}
\dfrac{
    \text{\# groups with all incorrect variants}
}{
    \text{\# total problem groups}
}
\end{array}
\]
\endgroup

\subsection{Results}

Table~\ref{tab:combined_metrics} provides a comprehensive evaluation of large language models on \BenchmarkName. Three models emerge as clear leaders: Anthropic Sonnet-3.7, Gemini-2.0-Flash~\cite{claude2024, gemini2024}, and Llama4-Maverick-17B-128E achieve Exact Match Accuracy exceeding 64\% with correspondingly high Partial Accuracy, demonstrating reliable multi-step reasoning capabilities.

Among the top performers, Sonnet-3.7 distinguishes itself with the highest Consistency Score (42.42\%) and lowest Confusion Rate (6.06\%), demonstrating superior robustness to paraphrased and perturbed problem variants, and Gemini-2.0-Flash achieves the highest Partial Accuracy (71.43\%). The Confusion Rate, which quantifies the proportion of problem groups where model accuracy across variants hovers around 50\%, reflects reasoning uncertainty; stronger models such as Sonnet-3.7 consistently display lower confusion rates. GPT-4-Turbo presents solid overall metrics (60.59\% partial, 53.73\% exact) but underperforms in consistency (33.33\%) compared to the top tier. Qwen2.5-72B-Instruct performs competitively (66.57\% partial, 61.69\% exact) but shows higher complete failure rates (15.09\%) than Maverick and Gemini. These results underscore the importance of evaluating mathematical reasoning not only in terms of accuracy, but also with respect to consistency, robustness, and failure modes, qualities that are often overlooked by traditional metrics, yet are essential for ensuring real-world reliability in scientific problem-solving.

Our in-depth analysis reveals that models such as Maverick are far more likely to succeed when guided by multiple-choice formats, particularly MC-Symbolic, compared to open-ended free-form questions. The structured nature of multiple-choice formats reduces the complexity of open-ended generation and enables models to focus on selecting the correct answer, which often leads to higher accuracy. While these models may still make conceptual errors, our results indicate that a substantial portion of their failures in free-form settings stem from challenges in generating complete and well-formatted solutions, as well as difficulties in arithmetic computation and physics-specific skills such as unit conversion. However, it is important to note that multiple-choice formats can provide additional cues or scaffolding that help the model arrive at the correct answer, even in the presence of partial understanding.
 Weaker models like 405B, however, show lower accuracy across all formats, suggesting more fundamental gaps in both conceptual understanding and execution. These findings underscore the diagnostic power of our benchmark. By systematically varying question formats, we are able to disentangle the underlying sources of model errors, yielding actionable insights for the advancement of scientific language models. Complete results, along with additional examples and extended analysis, are provided in Appendix~\ref{indepth_analysis}.



\section{Conclusion and Future Work}

We introduce \BenchmarkName, a benchmark for evaluating the scientific reasoning capabilities of large language models in physics, with a focus on generalization across diverse problem variations. Our benchmark assesses model robustness and introduces new metrics to capture output stability beyond standard accuracy.
Future work will expand \BenchmarkName\ to include multimodal reasoning tasks and interdisciplinary STEM domains, enabling the evaluation of models with complex, cross-domain scientific reasoning capabilities. This will drive the development of AI systems with robust, transparent, and reliable scientific reasoning.

\clearpage
\newpage
\bibliographystyle{assets/plainnat}
\bibliography{paper}

@String(AAAI = {AAAI})

@article{hendrycks2021measuring,
  title={Measuring mathematical problem solving with the math dataset},
  author={Hendrycks, Dan and Burns, Collin and Kadavath, Saurav and Arora, Akul and Basart, Steven and Tang, Eric and Song, Dawn and Steinhardt, Jacob},
  journal={arXiv preprint arXiv:2103.03874},
  year={2021}
}

@article{wang2023scibench,
  title={Scibench: Evaluating college-level scientific problem-solving abilities of large language models},
  author={Wang, Xiaoxuan and Hu, Ziniu and Lu, Pan and Zhu, Yanqiao and Zhang, Jieyu and Subramaniam, Satyen and Loomba, Arjun R and Zhang, Shichang and Sun, Yizhou and Wang, Wei},
  journal={arXiv preprint arXiv:2307.10635},
  year={2023}
}

@article{lu2022learn,
  title={Learn to explain: Multimodal reasoning via thought chains for science question answering},
  author={Lu, Pan and Mishra, Swaroop and Xia, Tanglin and Qiu, Liang and Chang, Kai-Wei and Zhu, Song-Chun and Tafjord, Oyvind and Clark, Peter and Kalyan, Ashwin},
  journal={Advances in Neural Information Processing Systems},
  volume={35},
  pages={2507--2521},
  year={2022}
}

@inproceedings{sun2024scieval,
  title={Scieval: A multi-level large language model evaluation benchmark for scientific research},
  author={Sun, Liangtai and Han, Yang and Zhao, Zihan and Ma, Da and Shen, Zhennan and Chen, Baocai and Chen, Lu and Yu, Kai},
  booktitle={Proceedings of the AAAI Conference on Artificial Intelligence},
  volume={38},
  number={17},
  pages={19053--19061},
  year={2024}
}

@article{kojima2022large,
  title={Large language models are zero-shot reasoners},
  author={Kojima, Takeshi and Gu, Shixiang Shane and Reid, Machel and Matsuo, Yutaka and Iwasawa, Yusuke},
  journal={Advances in neural information processing systems},
  volume={35},
  pages={22199--22213},
  year={2022}
}

@article{arora2023have,
  title={Have LLMs advanced enough? A challenging problem solving benchmark for large language models},
  author={Arora, Daman and Singh, Himanshu Gaurav and others},
  journal={arXiv preprint arXiv:2305.15074},
  year={2023}
}

@article{grattafiori2024llama,
  title={The llama 3 herd of models},
  author={Grattafiori, Aaron and Dubey, Abhimanyu and Jauhri, Abhinav and Pandey, Abhinav and Kadian, Abhishek and Al-Dahle, Ahmad and Letman, Aiesha and Mathur, Akhil and Schelten, Alan and Vaughan, Alex and others},
  journal={arXiv preprint arXiv:2407.21783},
  year={2024}
}

@article{bai2023qwen,
  title={Qwen technical report},
  author={Bai, Jinze and Bai, Shuai and Chu, Yunfei and Cui, Zeyu and Dang, Kai and Deng, Xiaodong and Fan, Yang and Ge, Wenbin and Han, Yu and Huang, Fei and others},
  journal={arXiv preprint arXiv:2309.16609},
  year={2023}
}

@article{jaiswal2024improving,
  title={Improving Physics Reasoning in Large Language Models Using Mixture of Refinement Agents},
  author={Jaiswal, Raj and Jain, Dhruv and Popat, Harsh Parimal and Anand, Avinash and Dharmadhikari, Abhishek and Marathe, Atharva and Shah, Rajiv Ratn},
  journal={arXiv preprint arXiv:2412.00821},
  year={2024}
}

@misc{gemini2024,
  author       = {{Google DeepMind}},
  title        = {Gemini},
  year         = {},
  howpublished = {\url{https://gemini.google.com/}},
  note         = {}
}

@misc{claude2024,
  author       = {{Anthropic}},
  title        = {Claude (Sonnet)},
  year         = {},
  howpublished = {\url{https://www.anthropic.com/claude/sonnet}},
  note         = {}
}

@article{zhu2024promptbench,
  title={Promptbench: A unified library for evaluation of large language models},
  author={Zhu, Kaijie and Zhao, Qinlin and Chen, Hao and Wang, Jindong and Xie, Xing},
  journal={Journal of Machine Learning Research},
  volume={25},
  number={254},
  pages={1--22},
  year={2024}
}

@article{bommasani2023holistic,
  title={Holistic evaluation of language models},
  author={Bommasani, Rishi and Liang, Percy and Lee, Tony},
  journal={Annals of the New York Academy of Sciences},
  volume={1525},
  number={1},
  pages={140--146},
  year={2023},
  publisher={Wiley Online Library}
}

@article{gardner2020evaluating,
  title={Evaluating models' local decision boundaries via contrast sets},
  author={Gardner, Matt and Artzi, Yoav and Basmova, Victoria and Berant, Jonathan and Bogin, Ben and Chen, Sihao and Dasigi, Pradeep and Dua, Dheeru and Elazar, Yanai and Gottumukkala, Ananth and others},
  journal={arXiv preprint arXiv:2004.02709},
  year={2020}
}

@article{larkin1979understanding,
  title={Understanding and teaching problem-solving in physics},
  author={Larkin, Jill H and Reif, F},
  journal={European journal of science education},
  volume={1},
  number={2},
  pages={191--203},
  year={1979},
  publisher={Taylor \& Francis}
}

@article{ahn2024large,
  title={Large language models for mathematical reasoning: Progresses and challenges},
  author={Ahn, Janice and Verma, Rishu and Lou, Renze and Liu, Di and Zhang, Rui and Yin, Wenpeng},
  journal={arXiv preprint arXiv:2402.00157},
  year={2024}
}

@article{lewkowycz2022solving,
  title={Solving quantitative reasoning problems with language models},
  author={Lewkowycz, Aitor and Andreassen, Anders and Dohan, David and Dyer, Ethan and Michalewski, Henryk and Ramasesh, Vinay and Slone, Ambrose and Anil, Cem and Schlag, Imanol and Gutman-Solo, Theo and others},
  journal={Advances in Neural Information Processing Systems},
  volume={35},
  pages={3843--3857},
  year={2022}
}

@article{chang2024survey,
  title={A survey on evaluation of large language models},
  author={Chang, Yupeng and Wang, Xu and Wang, Jindong and Wu, Yuan and Yang, Linyi and Zhu, Kaijie and Chen, Hao and Yi, Xiaoyuan and Wang, Cunxiang and Wang, Yidong and others},
  journal={ACM transactions on intelligent systems and technology},
  volume={15},
  number={3},
  pages={1--45},
  year={2024},
  publisher={ACM New York, NY}
}

@article{hegde2012they,
  title={How do they solve it? An insight into the learner’s approach to the mechanism<? format?> of physics problem solving},
  author={Hegde, Balasubrahmanya and Meera, BN},
  journal={Physical Review Special Topics—Physics Education Research},
  volume={8},
  number={1},
  pages={010109},
  year={2012},
  publisher={APS}
}

@article{reif1982knowledge,
  title={Knowledge structure and problem solving in physics},
  author={Reif, Frederick and Heller, Joan I},
  journal={Educational psychologist},
  volume={17},
  number={2},
  pages={102--127},
  year={1982},
  publisher={Taylor \& Francis}
}

@book{newell1972human,
  title={Human problem solving},
  author={Newell, Allen and Simon, Herbert Alexander and others},
  volume={104},
  number={9},
  year={1972},
  publisher={Prentice-hall Englewood Cliffs, NJ}
}

@article{10.7717/peerj-cs.103,
 title = {SymPy: symbolic computing in Python},
 author = {Meurer, Aaron and Smith, Christopher P. and Paprocki, Mateusz and \v{C}ert\'{i}k, Ond\v{r}ej and Kirpichev, Sergey B. and Rocklin, Matthew and Kumar, AMiT and Ivanov, Sergiu and Moore, Jason K. and Singh, Sartaj and Rathnayake, Thilina and Vig, Sean and Granger, Brian E. and Muller, Richard P. and Bonazzi, Francesco and Gupta, Harsh and Vats, Shivam and Johansson, Fredrik and Pedregosa, Fabian and Curry, Matthew J. and Terrel, Andy R. and Rou\v{c}ka, \v{S}t\v{e}p\'{a}n and Saboo, Ashutosh and Fernando, Isuru and Kulal, Sumith and Cimrman, Robert and Scopatz, Anthony},
 year = 2017,
 month = jan,
 volume = 3,
 pages = {e103},
 journal = {PeerJ Computer Science},
 issn = {2376-5992},
 url = {https://doi.org/10.7717/peerj-cs.103},
 doi = {10.7717/peerj-cs.103}
}

@misc{pint-library,
  title        = {Pint: Define, operate, and manipulate physical quantities},
  howpublished = {\url{https://pint.readthedocs.io/}},
  year         = {2025},
  note         = {Accessed May 7, 2025},
}

\clearpage
\newpage
\beginappendix
\section{Detailed Model Performance Analysis}\label{indepth_analysis}

In addition to standard metric evaluation, we conducted an in-depth analysis of model performance across multiple dimensions. For this analysis, we evaluate three model configurations: \texttt{llama4-maverick-17b-128e} (Maverick), \texttt{llama4-scout-17b-16e} (Scout), and \texttt{llama3.1-405b} (405B), presenting a comprehensive performance comparison.

\subsubsection{Performance by Textual Variant}

We analyze model performance across three textual variants, which differ in their surface phrasing while preserving the underlying question. Table~\ref{tab:template_performance} presents a detailed breakdown.

\begin{table*}[t]
\centering
\small
\caption{Accuracy by textual variant across models. Each variant represents a different surface phrasing of the same underlying question.}
\label{tab:template_performance}
\begin{tabular}{llcc}
\toprule
\textbf{Model} & \textbf{Textual Variant} & \textbf{Partial Accuracy (\%)} & \textbf{Exact Match Accuracy (\%)} \\
\midrule
Llama4-maverick-17b-128e-instruct
& Variant I & 69.81 & 64.16  \\
& Variant II & 69.17 & 63.54  \\
& Variant III & 70.82 & 64.86 \\
\cmidrule(lr){1-4}
Llama4-scout-17b-16e-instruct
& Variant I & 55.81 & 49.56 \\
& Variant II & 57.15 & 50.57  \\
& Variant III & 56.44 & 50.33  \\
\cmidrule(lr){1-4}
Llama3.1-405b-instruct
& Variant I & 42.18 & 34.03  \\
& Variant II & 43.36 & 34.10  \\
& Variant III & 42.78 & 35.24 \\
\bottomrule
\end{tabular}
\end{table*}

All models exhibit consistent performance across templates, with minimal variation.

\subsubsection{Performance by Question Type}

Table~\ref{tab:question_type_performance} presents accuracy by question format: free-form, MC-Numerical, and MC-Symbolic.

\begin{table*}[t]
\centering
\small
\caption{Accuracy by question type across models.}
\label{tab:question_type_performance}
\begin{tabular}{llccc}
\toprule
\textbf{Model} & \textbf{Type} & \textbf{Partial Accuracy (\%)} & \textbf{Exact Match Accuracy (\%)} & \textbf{Data\%} \\
\midrule
Llama4-maverick-17b-128e-instruct
& Free-Form & 65.72 & 57.69 & 71.52 \\
& MC Numerical & 65.23 & 65.23 & 14.24 \\
& MC Symbolic & 95.70 & 95.70 & 14.24 \\
\cmidrule(lr){1-5}
Llama4-scout-17b-16e-instruct
& Free-Form & 53.28 & 44.44 & 71.52 \\
& MC Numerical & 47.56 & 47.56 & 14.24 \\
& MC Symbolic & 81.51 & 81.51 & 14.24 \\
\cmidrule(lr){1-5}
Llama3.1-405b-instruct
& Free-Form & 36.61 & 24.95 & 71.52 \\
& MC Numerical & 59.42 & 59.42 & 14.24 \\
& MC Symbolic & 57.21 & 57.21 & 14.24 \\
\bottomrule
\end{tabular}
\end{table*}

A striking divergence emerges across models and question types. Maverick and Scout excel at MC-Symbolic questions (95.70\% and 81.51\%, respectively), while showing substantially lower performance on MC-Numerical questions. Manual analysis of 100 randomly sampled examples revealed that the performance gap between MC-Symbolic and MC-Numerical stems primarily from errors in numerical computation and unit conversion, rather than conceptual understanding deficits.

Free-form questions present a fundamentally different challenge and exhibit consistently lower performance across all models. This gap is attributable to two key factors: (1) structural complexity: free-form questions contain an average of two to three interconnected sub-questions that must be solved sequentially, with errors in early steps propagating to subsequent parts; and (2) evaluation stringency: models must generate complete, correctly formatted solutions rather than simply selecting from provided options. This combination of increased reasoning depth and answer generation requirements makes free-form questions substantially more demanding than their multiple-choice counterparts.

Surprisingly, 405B exhibits relatively balanced multiple-choice performance (59.42\% MC-Numerical, 57.21\% MC-Symbolic) but dramatically lower free-form accuracy (24.95\%).

\subsubsection{Performance Across Response Iterations}

We examine model stability across five response iterations to assess consistency. Table~\ref{tab:iteration_performance} summarizes the results. All models demonstrate stable performance across iterations.

\begin{table*}[t]
\centering
\small
\caption{Accuracy by iteration for all models. All values are percentages.}
\label{tab:iteration_performance}
\begin{tabular}{llcc}
\toprule
\textbf{Model} & \textbf{Iteration} & \textbf{Partial Accuracy (\%)} & \textbf{Exact Match Accuracy (\%)} \\
\midrule
Llama4-maverick-17b-128e-instruct
& 0 & 70.11 & 64.65 \\
& 1 & 70.29 & 64.24 \\
& 2 & 70.27 & 64.40 \\
& 3 & 69.49 & 63.99 \\
& 4 & 69.44 & 63.58 \\
\cmidrule(lr){1-4}
Llama4-scout-17b-16e-instruct
& 0 & 56.89 & 50.33 \\
& 1 & 56.96 & 51.16 \\
& 2 & 56.74 & 50.33 \\
& 3 & 56.28 & 49.67 \\
& 4 & 55.56 & 49.34 \\
\cmidrule(lr){1-4}
Llama3.1-405b-instruct
& 0 & 42.78 & 34.69 \\
& 1 & 42.75 & 34.27 \\
& 2 & 43.03 & 34.85 \\
& 3 & 42.77 & 34.19 \\
& 4 & 42.63 & 34.27 \\
\bottomrule
\end{tabular}
\end{table*}

\subsubsection{Cross-Type Error Analysis}

To investigate whether errors are question-format-specific or stem from deeper conceptual misunderstandings, we conducted a cross-type analysis on a subset of problems that appear in all three formats (free-form, MC-Numerical, and MC-Symbolic). This allows us to examine whether difficulty in one format predicts difficulty in others, revealing the nature of model errors.

\paragraph{Conditional Accuracy Analysis}

A critical question is whether model errors reflect format-specific challenges (e.g., numerical computation, answer generation) or fundamental conceptual misunderstandings. To distinguish these error types, we compute \emph{conditional accuracy}: for problems where a model fails in one format, what is its accuracy on the same problem presented in a different format?

If errors were primarily conceptual, models should fail consistently across all formats of the same problem, yielding low conditional accuracy. Conversely, high conditional accuracy indicates that the model understands the concept but fails due to format-specific requirements. Table~\ref{tab:conditional_accuracy} presents this analysis.

\begin{table*}[t]
\centering
\small
\caption{Conditional accuracy: given failure on one format (Condition), what is the average accuracy on another format (Target) for the same problems? High values indicate format-specific rather than conceptual errors. All values are percentages.}
\label{tab:conditional_accuracy}
\begin{tabular}{llcc}
\toprule
\textbf{Model} & \textbf{Given Failure On} & \textbf{Accuracy On} & \textbf{Success Rate (\%)} \\
\midrule
Llama4-maverick-17b-128e-instruct
& Free-Form & MC-Symbolic & 95.45 \\
& Free-Form & MC-Numerical & 60.18 \\
& MC-Numerical & MC-Symbolic & 95.00 \\
\cmidrule(lr){1-4}
Llama4-scout-17b-16e-instruct
& Free-Form & MC-Symbolic & 81.25 \\
& Free-Form & MC-Numerical & 46.33 \\
& MC-Numerical & MC-Symbolic & 79.55 \\
\cmidrule(lr){1-4}
Llama3.1-405b-instruct
& Free-Form & MC-Symbolic & 64.48 \\
& Free-Form & MC-Numerical & 54.06 \\
& MC-Numerical & MC-Symbolic & 60.93 \\
\bottomrule
\end{tabular}
\end{table*}

\paragraph{Interpretation and Key Insights}

The conditional accuracy patterns reveal fundamentally different error sources across models:

\begin{itemize}
    \item \textbf{MC-Numerical vs. MC-Symbolic Gap:} We assess cases where models fail on MC-Numerical questions and evaluate their accuracy on the corresponding MC-Symbolic versions. All models show substantially higher conditional accuracy on MC-Symbolic (95.00\%, 79.55\%, 60.93\%), indicating that most MC-Numerical errors are due to computational issues (e.g., arithmetic mistakes, unit conversion), rather than misunderstanding the underlying concepts or formulas. When explicit computation is removed, model performance improves markedly.

\item \textbf{Free-Form vs. Multiple-Choice Gap:} We examine cases where models fail on free-form questions and measure their accuracy on the corresponding MC-Numerical and MC-Symbolic formats. For example, Maverick achieves 95.45\% accuracy on MC-Symbolic and 60.18\% on MC-Numerical for problems it fails in free-form. This substantial improvement in accuracy with guided formats suggests that many free-form errors may stem from challenges in generating complete and well-formatted solutions, rather than from fundamental conceptual or arithmetic misunderstandings. However, it is important to note that multiple-choice formats can provide additional cues or scaffolding that help the model arrive at the correct answer, even in the presence of partial understanding. Thus, while the observed gap is indicative of generation and formatting challenges, it does not fully rule out the possibility of underlying conceptual gaps. Further analysis is needed to disentangle these effects.

\item \textbf{Model Differences:} Scout exhibits a similar but slightly weaker pattern, with conditional accuracies of 81.25\% (MC-Symbolic given free-form failure) and 79.55\% (MC-Symbolic given MC-Numerical failure). This indicates that most errors are still format-specific, though some conceptual gaps may remain. As with Maverick, the improvement in accuracy with guided formats highlights the benefit of scaffolding and structured problem presentation for model performance.
    \item \textbf{Conceptual Inconsistency in 405B:} In contrast, 405B shows lower conditional accuracies (60--65\%), indicating that a significant portion of its errors are due to inconsistent reasoning or incomplete conceptual understanding, even when the format is simplified.
\end{itemize}
\noindent\textbf{Implications for Model Development:} These findings suggest different improvement strategies for different model capabilities. For Maverick and Scout, gains would come from better solution generation, numerical precision, and unit handling, as their conceptual reasoning is already strong. For 405B, improvements require addressing fundamental reasoning consistency before tackling format-specific issues. The conditional accuracy metric thus serves as a diagnostic tool, revealing whether a model needs better conceptual understanding or improved execution.

\section{Distribution of Sub-Questions}\label{distribution sub-questions}

\begin{figure}[h]
    \centering
    \includegraphics[width=0.5\linewidth]{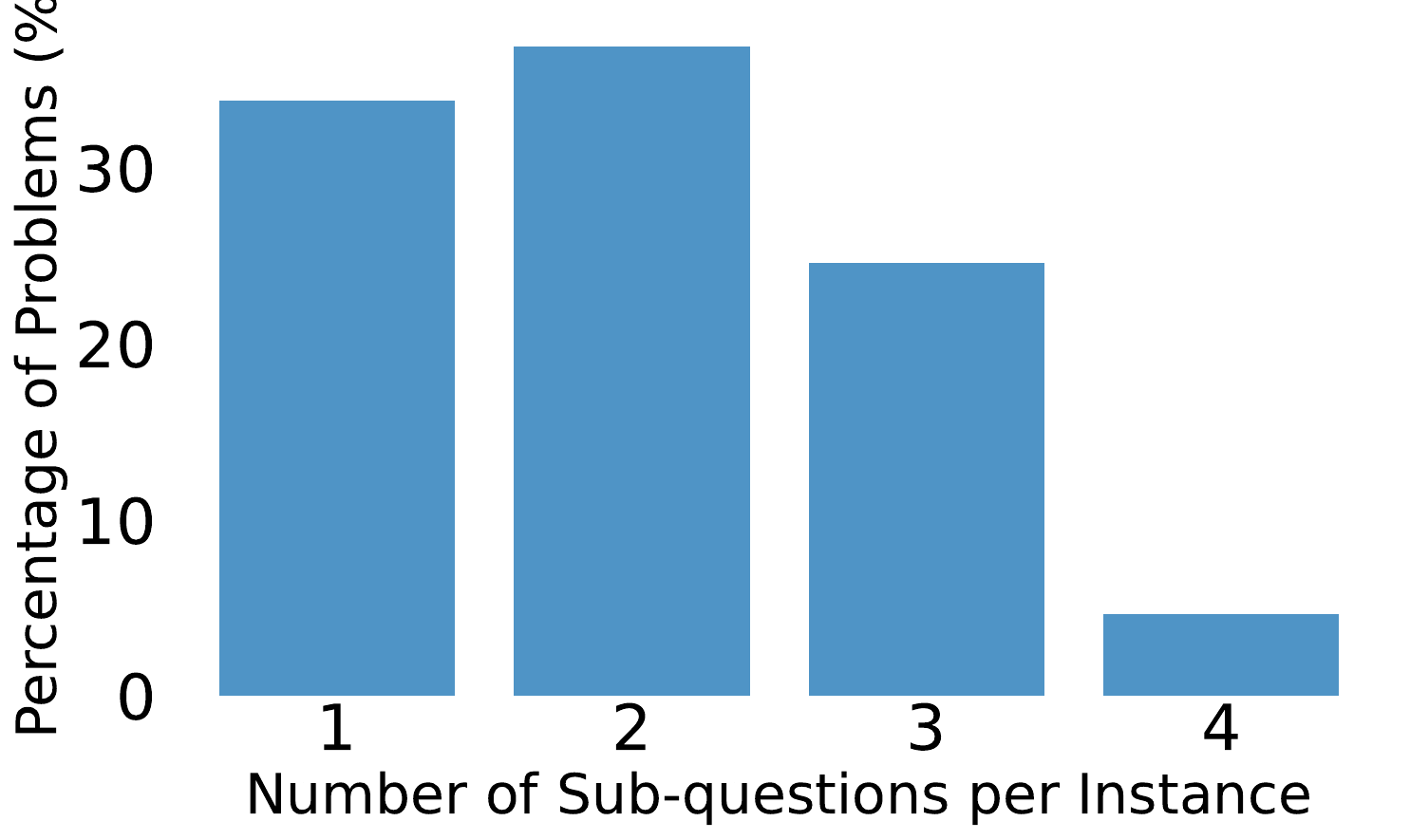}
    \caption{Distribution of \BenchmarkName~problems by number of sub-questions per instance.}
    \label{fig:subquestion_distribution}
\end{figure}

Real-world physics problems often consist of multiple interconnected sub-questions that build upon each other, requiring students to demonstrate cumulative understanding. To reflect this complexity, many problems in \BenchmarkName~are structured as multi-part questions. Figure~\ref{fig:subquestion_distribution} shows the distribution of problems across different numbers of sub-questions per instance.

\subsection{Example-Level Insights}

\begin{figure*}[t]
    \centering
    \includegraphics[width=15cm]{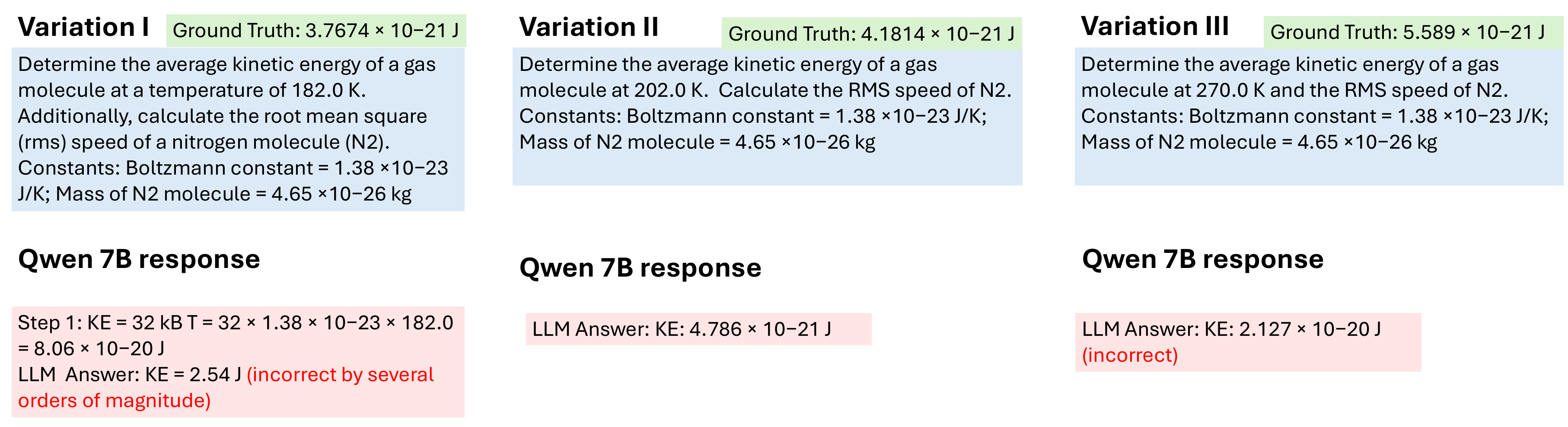}
    \caption{Three variation of the same question with different input variables. Only the Qwen-7B model's final step responses are shown.}
    \label{fig:example2}
\end{figure*}

While aggregate metrics provide a high-level view of model performance, deeper insights emerge when examining specific examples and their variations. In this section, we analyze representative cases that highlight recurring success patterns, consistent failure modes, and surprising behaviors across paraphrased or perturbed inputs.

These quantitative and qualitative observations shed light on the limitations of current models in terms of robustness, generalization, and interpretability.

\vspace{0.5em}
\noindent\textbf{Case Study I: Sensitivity to Input Variations}

We analyze three semantically equivalent versions of a physics question requiring symbolic reasoning and precise numerical calculation as shown in Figure \ref{fig:example2}. The first part of the task is to compute the average kinetic energy of a gas molecule nitrogen molecules at a given temperature. We analyze the result of Qwen2.5-7B model.

As shown in Figure \ref{fig:example2}, in the first variation, the model's answer is off by several orders of magnitude, indicating a fundamental flaw in its solution strategy. In the second variation, the response is approximately correct aside from a minor numerical discrepancy, reflecting improved accuracy under this paraphrased input. In the third variation, the kinetic energy is overestimated by more than three times, likely due to an arithmetic error or a misinterpretation of symbolic expressions. Without \BenchmarkName, such nuanced insights into model behavior that emerge from the same underlying question with different input realizations would remain inaccessible, limiting our ability to diagnose failure modes and evaluate robustness.

\vspace{0.8em}

Our observation is not limited to smaller models like Qwen2.5-7B. Even Qwen2.5-72B, despite using the correct physics formula and providing step-by-step reasoning, often produces numerically inconsistent results.

Consider the following example:

\begin{tcolorbox}[colback=white, colframe=mypink, title=Qwen2.5-72B: Incorrect Electric Field Calculatio, boxrule=1.25pt]
\textbf{Question:} Determine the magnitude of the electric field \( E \) generated by a point charge of \( 2.09 \times 10^{-9} \, \text{C} \) at a distance of \( 0.00567 \, \text{m} \). Use Coulomb’s constant \( k = 8.99 \times 10^9 \, \text{N} \cdot \text{m}^2/\text{C}^2 \).\\[0.5em]
\textbf{Qwen2.5-72B Answer:} Applies \( E = \frac{kq}{r^2} \), computes \textcolor{red}{\( E \approx 587.5 \, \text{N/C} \)}.\\[0.5em]
\textbf{Ground Truth:} \( E \approx 584{,}440 \, \text{N/C} \)
\end{tcolorbox}

\noindent The model uses the correct formula and walks through intermediate steps, but its final numeric output is off by nearly three orders of magnitude. This suggests that the issue is not conceptual misunderstanding but internal instability in arithmetic or symbolic execution. Similar inconsistencies were observed across other problems with slight input perturbations.

\vspace{0.5em}
\noindent\textbf{Case Study II: Measuring Hallucination}

A key strength of our benchmark is its ability to systematically induce and detect hallucinations in LLM model responses by dynamically altering or concealing critical components of a physics problem, such as input variables or domain-specific constants. This functionality enables controlled testing of model behavior under uncertainty. When confronted with incomplete information: \textit{Does the model seek clarification?} \textit{Does it make reasonable assumptions and state them explicitly?} \textit{Does it hallucinate values and proceed as if the input were fully specified?} Our benchmark is uniquely designed to probe these behaviors, allowing us to quantify reasoning integrity in under-specified scenarios and assess the robustness of models under challenging conditions.

We illustrate this with a representative example:

\begin{tcolorbox}[colback=white, colframe=teal!80!blue!40, title=Example of Hallucination, boxrule=1.25pt]
\textbf{Question:} Determine the average time required for a glucose molecule to diffuse a distance of 0.00991 m in water.
\end{tcolorbox}

This question omits the diffusion coefficient \(D\), a necessary constant for computing the answer via the physical equation \(t = \frac{x^2}{2D}\). However, when prompted with this version, Gemini does not request the missing value or flag the input as incomplete. Instead, the model fabricates a response by assuming an image is provided, stating: `Based on the image,' and then extrapolates from a hallucinated example involving diffusion over 0.010 m in \(7.5 \times 10^4\) seconds. It uses the proportionality \(t \propto x^2\) to calculate:
{
\[
t_2 = t_1 \cdot \left(\frac{x_2^2}{x_1^2}\right),
\]
\text{with}
{\small
\[t_1 = 7.5 \times 10^4 \, \text{s}, \, x_1 =0.010 \, \text{m}, \, x_2 = 0.00991 \, \text{m}
\]
}
}

leading to an incorrect answer.

This response reflects a hallucinated reasoning chain. Instead of applying the correct physics or querying for \(D\), the model infers a scenario that was never presented. Such behavior can be quantitatively evaluated in our benchmark by selectively omitting critical variables and analyzing how often models hallucinate versus recognize under-specified inputs.

In future work, we plan to formalize this capability and systematically benchmark hallucination rates across model families. This expands the scope of our dataset beyond correctness and robustness, making it a valuable tool for studying reasoning integrity under \textit{partial} or \textit{ambiguous} inputs.

\vspace{0.5em}
\noindent\textbf{Case Study III: Implicit Simplification Bias}

Another class of error arises in problems requiring more advanced topics. For example in question related to relativistic mechanics even when the scenario clearly demands relativistic treatment, the Qwen2.5-72B frequently defaults to oversimplified Newtonian expressions, for example using \( a = \frac{F}{m} \) or \( a = \frac{F}{\gamma m} \) without accounting for the orientation of the force relative to velocity. We call this behavior implicit simplification bias where the model superficially identifies relevant physical variables but fails to apply the correct governing equations when deeper conceptual distinctions are required. This suggests that such biases are not merely a consequence of model size but rather reflect fundamental gaps in their understanding of domain-specific complexities, highlighting the need for explicit training in these advanced areas.

\section{Error in Each Step in the Pipeline}
All problems were manually reviewed to ensure correctness. Table~\ref{tab:data_filtering} shows the error rate at each stage of the pipeline.

\begin{table*}[h]
\centering
\caption{Data Filtered Due to Errors at Each Stage of the Collection and Processing Pipeline}
\label{tab:data_filtering}
\resizebox{\textwidth}{!}{%
\begin{tabular}{lllp{6cm}}
\toprule
\textbf{Stage} & \textbf{Type of Error Checked} & \textbf{Error Rate (Percentage of data filtered)} \\
\midrule
1. Filtered Data &  Dependency to previous questions or visual information & $\sim$5\% (manually checked, incorrect ones filtered) \\
2. Structured Representation & Incorrect JSON structure & $\sim$4.5\% \\
3. Template Generation & Variable mismatch with Stage 3; Incorrect JSON structure & $\sim$1\% \\
4. Generating Variations & Incorrect JSON structure & $<$1\% \\
5. Python Code & Function signature mismatch; incorrect output; unit errors & $\sim$12\% \\
6. Final Manual Review & Human inspection & All remaining verified \\
\bottomrule
\end{tabular}%
}
\end{table*}

\section{Comparison with Existing Scientific Reasoning Benchmarks}
\label{appendix:comparison}

To illustrate the limitations of current scientific reasoning benchmarks, we provide representative examples from ScienceQA~\cite{lu2022learn} and SciEval~\cite{sun2024scieval}. These datasets primarily focus on selecting the correct answer from multiple choices, without requiring explicit, step-by-step reasoning or handling parameterized problem variations.

For example, consider the ScienceQA dataset:

\begin{graylisting}
"question": "Select the solid."
"choices": ["rain", "water in a fishbowl", "hammer"]
"answer": 2
\end{graylisting}

Or the SciEval benchmark:

\begin{graylisting}
"question": "How can momentum be decreased?"
"choices": [
  "A. Decrease mass or velocity, or transfer momentum through collision.",
  "B. Keep mass and velocity constant, avoid collisions.",
  "C. Increase mass and velocity, avoid collisions.",
  "D. Increase mass, decrease velocity, and avoid collisions."
]
"answer": ["A"]
\end{graylisting}

These examples highlight that most existing benchmarks emphasize answer selection rather than structured reasoning and do not support systematic variations in numerical parameters or textual formulations.

Table~\ref{tab:benchmark_comparison} provides a high-level comparison of existing physics benchmarks, illustrating how \BenchmarkName~addresses these limitations.

\begin{table*}[h]
\centering
\begin{adjustbox}{max width=\textwidth}
\renewcommand{\arraystretch}{1.1}
\begin{tabular}{>{\centering\arraybackslash}m{6.5cm} >{\centering\arraybackslash}m{1.5cm} >{\centering\arraybackslash}m{4cm} >{\centering\arraybackslash}m{2cm} >{\centering\arraybackslash}m{2cm} >{\centering\arraybackslash}m{2cm} >{\centering\arraybackslash}m{1cm} >{\centering\arraybackslash}m{2cm}}
\toprule
\textbf{Dataset} & \textbf{Number of Problems} & \textbf{Academic Level} & \textbf{Step-by-step Reasoning} & \textbf{Numerical Variation Q\&A} & \textbf{Textual Variation Q\&A} & \textbf{Python Code} & \textbf{Unit Validation} \\
\midrule
ScienceQA~\cite{lu2022learn} & 4,546 & Elem. \& Highschool & \xmarkred & \xmarkred & \xmarkred & \xmarkred & \xmarkred \\
SciBench~\cite{wang2023scibench} & 594 & Highschool & \xmarkred & \xmarkred & \xmarkred & \xmarkred & \xmarkred \\
SciEval~\cite{sun2024scieval} & 1,657 & Mixed & \xmarkred & Partial & \xmarkred & \xmarkred & \xmarkred \\
JEEBench~\cite{arora2023have} & 512 & Highschool & \xmarkred & \xmarkred & \xmarkred & \xmarkred & \xmarkred \\
MMLU Physics~\cite{hendrycks2021measuring} & 124 & Highschool & \xmarkred & \xmarkred & \xmarkred & \xmarkred & \xmarkred \\
PhysicsQA~\cite{jaiswal2024improving} & 370 & Highschool & \cmarkgreen & \xmarkred & \xmarkred & \xmarkred & \xmarkred \\
\midrule
\rowcolor[HTML]{E8F4FD}
\textbf{\BenchmarkName~(Ours)} & \textbf{\BenchmarkLength} & Undergraduate & \cmarkgreen & \cmarkgreen & \cmarkgreen & \cmarkgreen & \cmarkgreen \\
\bottomrule
\end{tabular}
\end{adjustbox}
\caption{\textbf{High-level comparison of physics benchmarks.} `Partial' indicates limited (2--3 variations) or inconsistent support.}
\label{tab:benchmark_comparison}
\end{table*}

\section{Examples from \BenchmarkName}\label{Moreexamples}
Here are some examples from \BenchmarkName, where each question is shown along with step-by-step reasoning and the corresponding Python code. We populate the problems with numerical values, and the relevant variables are generated as part of the pipeline.

\subsection{Example A}
\begin{tcolorbox}[colback=white, colframe=mypink, arc=1mm, title=Question, coltitle=white, fonttitle=\footnotesize\bfseries, boxrule=0.5pt, left=2pt, right=2pt, top=2pt, bottom=2pt, boxsep=2pt, fontupper=\fontsize{8.5pt}{9pt}\selectfont]

In a simplified atomic model, the most probable distance between the nucleus and an electron is \( r = 3.33e-11 \, \text{m} \). The nucleus contains \( 1.3 \) protons. Determine the electric field due to the nucleus at the electron's position.

Here are constants:
\[
\text{Permittivity of free space} = 8.85 \times 10^{-12} \, \frac{\text{C}^2}{\text{N} \cdot \text{m}^2}
\]
\[
\text{Elementary charge} = 1.6 \times 10^{-19} \, \text{C}
\]
\end{tcolorbox}

\begin{tcolorbox}[colback=white, colframe=mygreen, arc=1mm, title=Solution, coltitle=white, fonttitle=\footnotesize\bfseries, boxrule=0.5pt, left=2pt, right=2pt, top=2pt, bottom=2pt, boxsep=2pt, fontupper=\fontsize{8.5pt}{9pt}\selectfont]

\textbf{Identify Relevant Concepts}
\begin{itemize}
    \item The electric field due to a point charge is given by
    \[
    \vec{E} = \frac{1}{4\pi\epsilon_0} \frac{q}{r^2} \hat{r}
    \]
    where \( \epsilon_0 \) is the permittivity of free space, \( q \) is the charge, and \( r \) is the distance from the charge.

    \item The goal is to calculate the electric field at the electron's position.
\end{itemize}

\textbf{Set Up the Problem}
\begin{itemize}
    \item The electric field at a distance \( r \) from a point charge is given by the formula above.
    \item The direction of the electric field is radially outward from the nucleus.
\end{itemize}

\textbf{Execute the Solution}
\begin{itemize}

    \item Substituting the given values into the formula
\end{itemize}

\textbf{Evaluate Your Answer}
\begin{itemize}
    \item The electric field is expected to be radially outward from the nucleus due to its positive charge.
    \item If \( r \) were very small, the electric field would be very large, and if \( r \) were large, the electric field would approach zero, which is physically reasonable.
\end{itemize}

\end{tcolorbox}

\begin{lstlisting}[
    language=Python,
    title={\small \textbf{Python code:}},
    label={lst:electric_field},
    basicstyle=\ttfamily\scriptsize, % Smaller font size
    keywordstyle=\color{blue},
    commentstyle=\color{green!60!black},
    breaklines=true,
    columns=flexible,
    literate={µ}{$\upmu$}1
        {τ}{$\uptau$}1
        {ω}{$\upomega$}1
        {y_m}{$y_{m}$}1
        {P_avg}{$P_{\text{avg}}$}1]
import sympy as sp
from pint import UnitRegistry
ureg = UnitRegistry()
Q_ = ureg.Quantity

def electric_field_at_electron(r, e, number_of_protons, epsilon_0):
    # Convert inputs to Pint quantities
    r = Q_(r).to(ureg.meter)  # Ensure meters
    e = Q_(e).to(ureg.coulomb)  # Ensure coulombs
    number_of_protons = Q_(number_of_protons).to(ureg.dimensionless)
    epsilon_0 = Q_(epsilon_0).to(ureg.farad / ureg.meter)  # Ensure F/m

    r = r.magnitude
    e = e.magnitude
    number_of_protons = number_of_protons.magnitude
    epsilon_0 = epsilon_0.magnitude

    # Define symbolic variables
    q = e * number_of_protons
    E = sp.Symbol('E', real=True, positive=True)

    # Calculate the electric field
    E = (1 / (4 * sp.pi * epsilon_0)) * (q / r**2)

    return {
        'E': E.evalf()
    }

\end{lstlisting}

\newpage

\subsection{Example B}
\begin{tcolorbox}[colback=white, colframe=mypink, arc=1mm, title=Question, coltitle=white, fonttitle=\footnotesize\bfseries, boxrule=0.5pt, left=2pt, right=2pt, top=2pt, bottom=2pt, boxsep=2pt, fontupper=\fontsize{8.5pt}{9pt}\selectfont]

Consider a solid metal cube with an edge length of \( L = 0.0237 \, \text{m} \).

(a) Determine the lowest energy level for an electron within this metal.

(b) Calculate the energy difference between this level and the next higher energy level.

Here are constants:
\begin{align*}
\text{Reduced Planck's constant } \hbar &= 1.05 \times 10^{-34} \, \text{J} \cdot \text{s} \\
\text{Electron mass } m_e &= 9.11 \times 10^{-31} \, \text{kg} \\
\text{Ground state quantum numbers: } &\quad n_x = n_y = n_z = 1 \\
\text{Next state quantum numbers: } &\quad n_x = 2, \, n_y = 1, \, n_z = 1
\end{align*}

\end{tcolorbox}

\begin{tcolorbox}[colback=white, colframe=mygreen, arc=1mm, title=Solution, coltitle=white, fonttitle=\footnotesize\bfseries, boxrule=0.5pt, left=2pt, right=2pt, top=2pt, bottom=2pt, boxsep=2pt, fontupper=\fontsize{8.5pt}{9pt}\selectfont]

\textbf{Identify Relevant Concepts}
\begin{itemize}
    \item Model the electron as a particle in a 3D box.
    \item Energy levels are given by:
    \[
    E(n_x, n_y, n_z) = \frac{\pi^2 \hbar^2}{2m_e L^2}(n_x^2 + n_y^2 + n_z^2)
    \]
\end{itemize}

\textbf{Set Up the Problem}
\begin{itemize}
    \item Ground state: \( n_x = n_y = n_z = 1 \)
    \item Next higher level: \( n_x = 2, n_y = 1, n_z = 1 \)
\end{itemize}

\textbf{Execute the Solution}
\begin{itemize}
    \item Compute:
    \[
    E_1 = \frac{\pi^2 \hbar^2}{2m_e L^2}(1^2 + 1^2 + 1^2)
    \]
    \[
    E_2 = \frac{\pi^2 \hbar^2}{2m_e L^2}(2^2 + 1^2 + 1^2)
    \]
    \item Energy difference:
    \[
    \Delta E = E_2 - E_1
    \]
\end{itemize}

\textbf{Evaluate Your Answer}
\begin{itemize}
    \item Positive energy difference is expected since next level is higher.
    \item Larger cube size would reduce energy spacing, consistent with quantum model.
\end{itemize}

\end{tcolorbox}

\begin{lstlisting}[
    language=Python,
    title={\small \textbf{Python code:}},
    label={lst:energy_levels},
    basicstyle=\ttfamily\scriptsize,
    keywordstyle=\color{blue},
    commentstyle=\color{green!60!black},
    breaklines=true,
    columns=flexible
]
import sympy as sp
from pint import UnitRegistry
ureg = UnitRegistry()
Q_ = ureg.Quantity

def electron_energy_levels(L, h_bar, m_e, n_x, n_y, n_z, n_x_next, n_y_next, n_z_next):
    L = Q_(L).to(ureg.meter)
    h_bar = Q_(h_bar).to(ureg.joule * ureg.second)
    m_e = Q_(m_e).to(ureg.kilogram)

    L = L.magnitude
    h_bar = h_bar.magnitude
    m_e = m_e.magnitude

    pi = sp.pi

    def energy(n_x, n_y, n_z, L, h_bar, m_e):
        return (pi**2 * h_bar**2 / (2 * m_e * L**2)) * (n_x**2 + n_y**2 + n_z**2)

    E1 = energy(n_x, n_y, n_z, L, h_bar, m_e)
    E2 = energy(n_x_next, n_y_next, n_z_next, L, h_bar, m_e)
    DeltaE = E2 - E1

    return {
        'E1': E1.evalf(),
        'E2': E2.evalf(),
        'DeltaE': DeltaE.evalf()
    }


\end{lstlisting}

\end{document}